# Impersonation: Modeling Persona in Smart Responses to Email

**Rajeev Gupta, Ranganath Kondapally, Chakrapani Ravi Kiran S**
Microsoft India AI & R
{rajgup, rakondap, ravichak} @microsoft.com

## Abstract

In this paper, we present design, implementation, and effectiveness of generating personalized suggestions for email replies (Smart Reply [Kannan *et al.*, 2016]). To personalize email responses based on user's style and personality, we model the user's persona based on her past responses to emails. This model is added to the language-based model created across users using past responses of the all user emails.

A user's model captures the typical responses of the user given a particular context. The context includes the email received, recipient of the email, and other external signals such as calendar activities, preferences, etc. The context along with user's personality (e.g., extrovert, formal, reserved, etc.) is used to suggest responses. These responses can be a mixture of multiple modes: email replies (textual), audio clips, etc. This helps in making responses mimic the user as much as possible and helps the user to be more productive while retaining her mark in the responses.

## 1 Introduction

Emails are an important means of communication in the digital arena. Though newer forms of communication such as Facebook and messaging applications have become dominant over email, it still is a primary means of communication for certain tasks. In particular, it is preferred for most of the formal interactions and hence, plays a major role in enterprise communications. Quite often, such communications are recurrent in nature and a lot of them have similar contexts. Even emails containing informal conversations, not a lot of them require cognitive effort from the user. Over time, the user tends to have a small set of standard responses to such emails. Popular email clients such as Outlook and Gmail provide suggestions to replies to help user save time and effort [Bullock *et al.*, 2017]. These suggestions are particularly helpful on mobile devices where typing can be tedious. In all the work proposed in literature, a single unified model is used to generate these suggestions for all the users. This model is trained on all the emails and their responses. The model takes the email text as input and generates a sequence of words as output. However different users reply to the same email differently based on their personality, mood, or the relationship with the person she is replying to. One standard response across users, oblivious to the persona of the user, takes away the personal touch from the responses and would be jarring experience to the recipient. Table 1 gives examples where people with different personalities *emotional*, *reserved* and *neutral* will respond to same email differently. In this paper, we present our ongoing work to generate such personalized responses.

A person's character reflects in her actions, demeanor, and words the person uses to communicate. In particular, in the digital world where predominantly words are the most visible one among the three aspects, her interactions naturally reflect her persona. In other words, her responses in various contexts to various people in written text captures her persona to a large extent. Some of the characteristics of a response to a given context that embody the persona are the following:
   a) *Emotion*: given any context, different people tend to react with different emotions. In order to impersonate the user, we need to understand and model this aspect in our responses. For example, a user may be dismissive about positive sentiments in the email but may be concerned about negative sentiments.
   b) *Degree of emotion*: It is not enough to model just the emotion of the response, but we need to capture the degree of the emotion as well. For example, a user might be very harsh on her team but soft on other teams: "We were *terrible* today" vs "They were *bad* today". Further, the degree of emotions varies depending on the recipient. For example, mails to personal relations contains more words showing emotions compared to mails to office colleagues.
   c) *Length of response*: Some people like to be terse or concise versus some people like to elaborate and are verbose. It is a part of their personality that reflects in the length of their responses.
   d) *Tone*: User preferences reflect partially in her responses. For example, a person may understand Machine learning but not Information retrieval. Her responses will likely have a tone of indifference in case of retrieval but will likely have a *concerned* and *informed* tone in case of ML.

Table 1: Different responses by different users

| Personalities | Emotion | Degree of emotion | Length | Tone |
|---|---|---|---|---|
| *Emotional* | "I am dying to meet you" "I am terribly sorry; I can't make it" | "I am free as a bird; Let's go" | "This is one hell of a news that is worth celebrating" | "This is mind blowing. I am approving this right away" |
| *Neutral* | "I would like to meet you" "Sorry; I can't make it" | "I am free; Let's go" | "Great to hear" | "Very good. Approved" |
| *Reserved* | "We should meet" "Sorry; I can't" | "Ok. Let's go" | "Nice!" | "Ok. Will see." |

In this paper, we stick to modeling the digital persona of the user. The digital persona may be different from the real-life persona of the user as people tend to be more conscious and watchful given memories in digital world tend to stay lot longer. Furthermore, we stick to modeling the persona that is visible from emails alone for the context of this paper. Our work can be extended in situations where the persona is modeled using various other inputs (e.g., browsing history, social media profile, etc.) besides user's emails and then used for replying emails.

## 2  Related Works

This work of automatically generating personalized replies has similarities in different parts with different works in literature. Authors of [Kannan *et al.*, 2016] give work closest to ours, describing smart-reply system of Gmail. They also give a method to generate a number of auto-reply suggestions which are semantically diverge. Authors cover the diversity and scalability issues of their neural-network-based architecture. A persona based neural conversation model is presented in [Jiwei *et al.*, 2016]. They generate embedding for each person based on her background information and speaking style. Authors present results showing improvements in perplexity and BLEU scores over simple sequence-to-sequence models. [Heudin, 2017] gives a method of emotion selection in conversational agents. Authors show that multiple personality conversational agent with appropriate emotion selection has better user-engagement compared to a neutral system. Authors of [Sherman *et al.*] present a neural network method to mimic the language of President Donald Trump. We are applying the similar method to understand individual user's email habits and mimicking its style to generate smart replies.

## 3  Architecture

The main task of our system is to generate a list of most likely responses. Figure 1 shows the architecture of our system. First, we use emails from all the users to create response creation model. Then we use the emails of one particular user to get the user specific model. Here are the different components of the architecture:

1. *Email classifier*: A user need not answer each email and even among the ones she needs to answer, some may not be suitable for automated answer generation. We train a binary classification model to identify emails that need an automatic response. The model is trained using logistic regression with L2 regularization and no bias. We preprocess each email to remove salutations, closing, and signatures. We also remove non-dictionary words from the email. We then create the following features to feed logistic regression classifier: *n-grams* (specifically, bigrams, trigrams, and quad-grams) from the email text and the length of the email (short, medium, and long). For generating ground truth for the same, we take a corpus of emails and tag ones whose response was either not given or whose response is longer than certain size as negatives and rest as positives. Specifically, if the email is responded back and the response size was two sentences or less (~25 words), we annotate such emails as the emails whose response should be attempted automatically. We ensure that the proportion of positives and negatives in the dataset is balanced. For the scope of this paper, we ignore some of the user experience related factors that are relevant to the decision of automatic response such as time of the day, type of device the email is being replied from, location of the user, etc.

2. *Request and response extractor*: This component has multiple parts. From the emails identified by the *Email Classifier*, we need to identify sentences in the incoming email, which need a reply. We use a LSTM-based sentence-classification model that classifies a sentence into request/question vs others (e.g., [Kim, 2014, Hassan *et al.*, 2017]). Briefly, our architecture for the model consists of an embedding layer that uses vectors of length 64 to represent each word, LSTM with 100 memory units, dense output layer with 1 neuron and sigmoid activation function to make the binary prediction. We use log loss as the loss function, batch size of 64, drop out probability of 0.2, and 3 epochs. Using the trained sentence classification model, we feed the email sentences and classify whether the sentence represents a request or question. Only the sentences, which represent request/question, are passed on to the next stage. If we do not identify any such sentence, we assume that the reply cannot be generated automatically, hence, rejected. For extracting responses, we consider the reply to the email as the response after standard preprocessing i.e., removing salutations, non-dictionary words, closing and signatures. Our dataset consists of 10K manually labelled sentences, extracted from twitter data, donated emails and consists of balanced proportions of questions, requests and others. We also leverage question classifiers dataset from Li, Roth [Li *et al.*, 2002]. For all users, we compute persona features from their responses such as average response length, emotion, degree of emotion (strong, mild, or neutral) by using sentiment analysis on the dataset. We use these features along with each request/question and response pair during training.

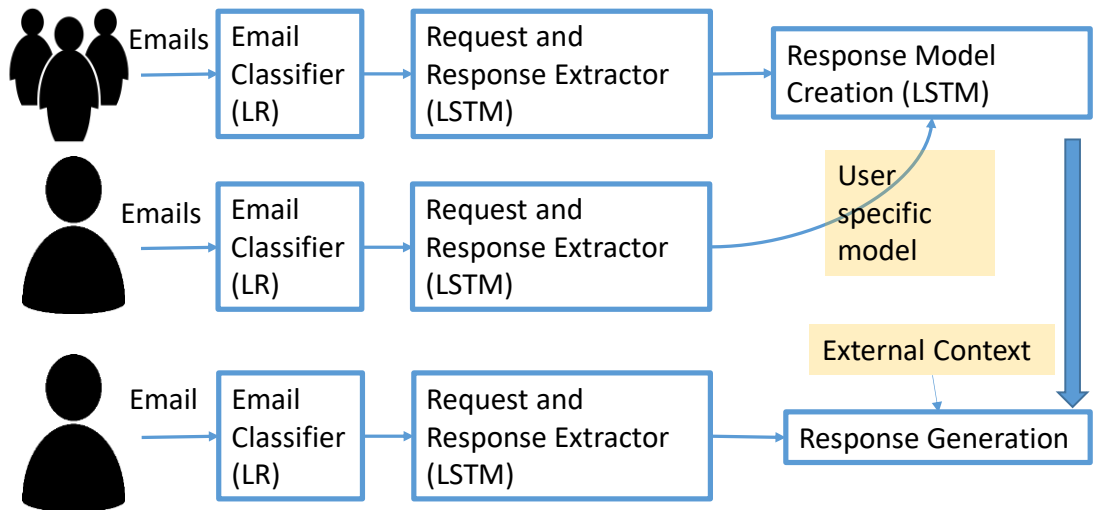

*Figure 1: System Block Diagram*

3. *Response Model Creation*: In this stage, we create a sequence-to-sequence smart reply model, which takes questions/requests from incoming email and outputs possible responses to them. We generate the training/test data for this stage using the previous stages on donated emails---we run the classifier built in stage 1 to filter the positives and then run the extractor built in stage 2 to identify the question/request and reply pairs from them---and twitter data. In total, our dataset consists of about 100K pairs, which we randomly split 80-20 into train and test sets. We then train a LSTM-based sequence-to-sequence model on the training data ([Sutskever *et al.*, 2014]). Our architecture consists of embedding layer that uses vectors of length 64, encoder LSTM layer with 256 memory cells, a decoder LSTM layer with 256 memory cells, a dense output layer, and a *softmax* layer. Note that the input to decoder LSTM is not the output of encoder LSTM but instead final hidden and cell states act as the initial state for the decoder. Similarly, the decoder passes its sequence of hidden states to the dense layer for output. After training the model, instead of generating one response, we modify the last layer to generate a number of responses along with their probabilities (removing *soft-max* layer).

4. *User Specific Model*: This step is required to personalize the smart reply model. Since individual user has small amount of data, we model the user based on *n-grams* used and map it to one of the users whose data was used in training. We boost the dataset size by finding similar questions/requests using semantic similarity techniques [Parikh *et al.*, 2016]. All similar questions/requests are assigned the same answer as given by the user. A rule-based engine is used to handle external signals. This engine takes responses generated by LSTM model as input and selects one depending on external signals. In this module we look for certain features in the responses (e.g., words like *meet*, *meeting*, *see you*, *day name*, etc.) and if those features are there, we take the calendar inputs to rank the results. Please see the appendix for our intended schematic.

5. *Response generation*: We extract the email features using request extractor and run the model to get the response. After getting multiple responses we use the rule-based module to rank the suggested responses.

## 4 Measurements

We want to measure how well the trained system mimics the responses of the user given a context and therefore, we use BLEU score [Papineni *et al*., 2002] between generated responses and the past responses of the user for a given email. We measure them across recipient types to verify diversity. We also compare the variance of generated responses for different users compared to generic automatic response generator. We continue to work with more training data and performance improvements along different metrics.

## 5 Conclusions and Future Work

We present design, architecture, and implementation of a personalized smart reply system. The aim of this system is to mimic the user as much as possible rather than user accepting some common denominator while selecting output of a sequence-to-sequence generator. We explained all the inputs required for the system, various neural components, and metrics, which will be used to measure the effectiveness of the system. As a part of our future work, we plan to improve the system by taking external signals as part of personalized training. Since we do not have calendar entries to train along with emails, we have separate rule-based system. We need to resolve these issues to have single integrated training and response system. We also need to consider the scenario where a user *behaves* differently while sending mails to different recipients. One can categorize recipients into various buckets (work, friends, relatives, etc.) [Culotta *et al*., 2004] to feed that as another feature to the user's model. All these are part of our future work.

**Appendix: Screenshots of output**

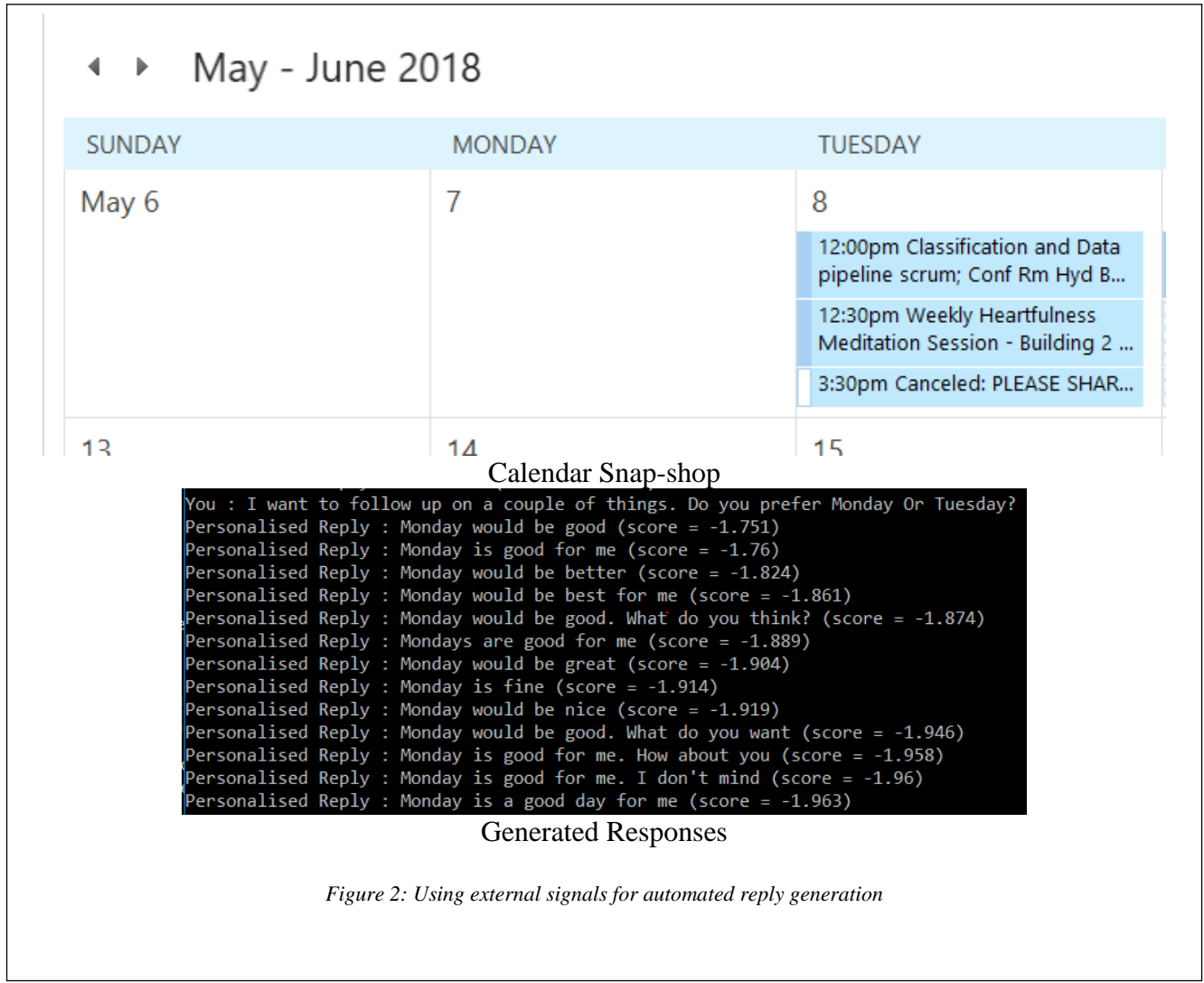

Calendar Snap-shop

Generated Responses

*Figure 2: Using external signals for automated reply generation*

This work was presented at 1st Workshop on Humanizing AI (HAI) at IJCAI'18 in Stockholm, Sweden